\documentclass{article}

\PassOptionsToPackage{numbers,compress}{natbib}
\usepackage[preprint]{neurips_2025}

\usepackage[utf8]{inputenc} % allow utf-8 input
\usepackage[T1]{fontenc}    % use 8-bit T1 fonts
\usepackage{hyperref}       % hyperlinks
\usepackage{url}            % simple URL typesetting
\usepackage{booktabs}       % professional-quality tables
\usepackage{amsfonts}       % blackboard math symbols
\usepackage{nicefrac}       % compact symbols for 1/2, etc.
\usepackage{microtype}      % microtypography
\usepackage{xcolor}         % colors
\usepackage{enumitem}
\usepackage{amsmath,amssymb,mathtools}
\usepackage{booktabs}
\usepackage{threeparttable}
\usepackage[table]{xcolor}

\usepackage{float}
\usepackage{booktabs}
\usepackage{caption}
\usepackage{graphicx}

\newcommand{\best}[1]{\textbf{#1}}
\newcommand{\second}[1]{\underline{#1}}

\title{Let the Abyss Stare Back\\[0.35em]
{\large Adaptive Falsification for Autonomous Scientific Discovery}}

\author{%
  Peiran Li \\
  Texas A\&M University\\
  VecTrue AI \\
  \texttt{lipeiran@tamu.edu}
  \And
  Fangzhou Lin\\
  Texas A\&M University\\
  Worcester Polytechnic Institute\\
  \And
  Shuo Xing\\
  Texas A\&M University
  \And 
  Jiashuo Sun\\
  University of Illinois Urbana-Champaign \\
  \And
  Dylan Zhang\\ 
  University of Illinois Urbana-Champain \\ 
  \And 
  Siyuan Yang\\
  Texas A\&M University\\
  \And
  Chaoqun Ni\\
  University of Wisconsin-Madison\\
  VecTrue AI\\
  \texttt{chaoqun.ni@wisc.edu}
   \And
  Zhengzhong Tu\\
  Texas A\&M University\\
  \texttt{tzz@tamu.edu}
}

\begin{document}

\maketitle

\begin{abstract}
Autonomous scientific discovery is entering a more dangerous regime: once the evaluator is frozen, a sufficiently strong search process can learn to win the exam without learning the mechanism the task was meant to reveal. This is the idea behind our title. To let the abyss stare back is to make evaluation actively push against the candidate through adaptive falsification, rather than passively certify it through static validation. We introduce DASES, a falsification-driven framework in which an Innovator, an Abyss Falsifier, and a Mechanistic Causal Extractor co-evolve executable scientific artifacts and scientifically admissible counterexample environments under a fixed scientific contract. In a controlled loss-discovery problem with a single editable locus, DASES rejects artifacts that static validation would have accepted, identifies the first candidate that survives the admissible falsification frontier, and discovers FNG-CE, a loss that transfers beyond the synthetic discovery environment and consistently outperforms CE and CE+L2 under controlled comparisons across standard benchmarks, including ImageNet.
\end{abstract}

\section{Introduction}
\label{sec:intro}

Since the emergence of large language models (LLMs)~\citep{llama3.2,yang2025qwen3,team2025gemma3,singh2025openai,pan2024plum,xing2025re,dai2025language,li2025safeflow,li2026bibagent,li2026traversal,lin2026position}, autonomous AI research has crossed a real threshold.
Systems can now automatically design agentic workflows, search program space for new algorithms and neural primitives, execute end-to-end research loops, and even produce manuscripts that pass peer review in workshop settings \cite{hu2024adas,wang2025evoagentx,lu2024aiscientist,yamada2025aiscientistv2,weng2025deepscientist,yu2025alpharesearch,pu2026pievo,vitvitskyi2026mining}.
This is a remarkable shift in capability.
But it also reveals a deeper methodological problem:
the principal bottleneck of automated scientific discovery is no longer merely insufficient search over hypotheses.
It is the continued reliance on \emph{static evaluation environments}.

A static evaluator is a frozen exam.
Once the exam is frozen, a sufficiently capable discovery system need not uncover the mechanism that truly governs the task; it only needs to learn how to score well on that exam.
It can exploit latent shortcuts, benchmark-specific regularities, proxy-lab artifacts, or evaluation loopholes while still appearing to make scientific progress.
This is the central danger of increasingly powerful autonomous research systems:
as the searcher becomes more capable, the evaluator becomes easier to overfit.
What looks like discovery may in fact be \emph{cognitive overfitting to the benchmark itself}.

Recent progress across autonomous research makes this tension impossible to ignore.
On the proposal side, systems for automated agent and workflow design show that AI can already invent increasingly strong research organizations and execution strategies \cite{hu2024adas,wang2025evoagentx}.
On the end-to-end side, AI Scientist-style systems and their successors now span hypothesis generation, experiment implementation, result analysis, figure production, and manuscript writing \cite{lu2024aiscientist,yamada2025aiscientistv2,weng2025deepscientist}.
On the discovery side, program-evolution and principle-evolution frameworks have expanded the search frontier from algorithms to broader scientific priors and open-ended problems \cite{yu2025alpharesearch,pu2026pievo,vitvitskyi2026mining}.
These advances establish, decisively, that autonomous systems can search.
But they also sharpen a shared limitation:
the candidate is still predominantly selected against a largely pre-specified evaluator---a benchmark suite, a proxy task family, a reward model, a review model, or a validation protocol chosen in advance \cite{hu2024adas,wang2025evoagentx,lu2024aiscientist,yamada2025aiscientistv2,weng2025deepscientist,yu2025alpharesearch,pu2026pievo,vitvitskyi2026mining}.

Evidence is now emerging that this is not a minor implementation issue, but a structural one.
In activation-function discovery, evolutionary search was shown to find candidates that exploit properties of a synthetic proxy lab, including batch-level statistical structure, producing impressive in-lab gains while sacrificing broader transferability and, in some cases, leading to pathological downstream behavior \cite{vitvitskyi2026mining}.
In parallel, recent work on statistical rigor in AI-driven discovery argues that without hard structural safeguards, autonomous research loops are vulnerable to spurious findings induced by sequential multiple testing, protocol leakage, and evaluator-facing shortcuts \cite{sargsyan2025structural}.
These observations point to the same conclusion:
\emph{if the evaluator does not adapt, the searcher will eventually learn the evaluator faster than it learns the science.}

This paper starts from a different premise.
If the aim of autonomous discovery is not merely to output candidates that score well, but to produce claims that deserve scientific belief, then discovery cannot remain organized as \emph{propose, score, retain} under a frozen evaluator.
It must be organized as \emph{propose, falsify, diagnose, and only then retain}.
A candidate should not be accepted because it performs well on a static benchmark.
It should be accepted because repeated, validity-preserving attempts to break it fail.
This requires elevating the evaluator itself into a first-class scientific object:
not an unconstrained attacker that injects arbitrary damage, but a generator of \emph{scientifically admissible} counterexamples that preserve task-defining semantics while probing shortcut dependence, brittle assumptions, tail-risk failure, and failures of mechanism-level generalization.

We introduce \textbf{DASES} -- \textbf{D}ynamic \textbf{A}dversarial \textbf{S}cientific \textbf{E}nvironment \textbf{S}ynthesis and Mechanistic Co-Evolution -- a framework built around exactly this shift.
DASES reframes automated discovery as a constrained scientific game between three roles.
An \emph{Innovator} proposes an executable scientific artifact through a \emph{single editable scientific locus}, ensuring that the object of scientific intervention remains explicit and isolated.
An \emph{Abyss Falsifier} performs \emph{dynamic adversarial environment synthesis}, constructing \emph{candidate-conditioned}, \emph{validity-constrained} falsification suites designed to expose hidden shortcut reliance and non-mechanistic success.
A \emph{Mechanistic Causal Extractor} converts the resulting diagnostic trace into an explicit account of \emph{why} the candidate failed or survived, and translates that account into a minimal next-round correction.
In DASES, the candidate artifact and the environments capable of refuting it do not sit on opposite sides of a fixed benchmark;
they \emph{co-evolve}.

This changes the object of optimization.
DASES does not seek the artifact that performs best on a frozen testbed.
It seeks the artifact that survives the strongest admissible falsification frontier currently available.
Invariance tests, counterfactual interventions, tail-regime stressors, and compositional adversaries are therefore not auxiliary diagnostics appended after the fact.
They are the engine of discovery itself.
This is the core methodological departure of our work:
DASES is not a stronger search heuristic, a more elaborate benchmark, or another autonomous scientist system.
It is a redefinition of what \emph{counts} as scientific progress for autonomous discovery.

A second principle is equally important.
Once evaluation becomes adaptive, scientific integrity cannot remain a post-hoc patch.
An adaptive evaluator raises epistemic pressure, but it also raises the stakes of protocol tampering, silent code drift, data contamination, and unsupported claims.
DASES therefore embeds a \emph{scientific contract} directly into the method:
a single authorized edit interface, an immutable base protocol outside that interface, isolated execution, replayable environment specifications, full provenance, and evidence-anchored external knowledge.
The result is not only a more robust artifact, but an auditable falsification trace explaining why weaker candidates did not deserve to survive.

\noindent\textbf{Contributions.}
Our contributions are fourfold:
\begin{itemize}[leftmargin=1.5em, itemsep=1pt, topsep=2pt, parsep=0pt, partopsep=0pt]
    \item We identify \emph{static evaluation environments} as a central, under-formalized bottleneck in autonomous scientific discovery, and frame the resulting failure mode as \emph{cognitive overfitting to the benchmark itself}.
    \item We introduce DASES, a general framework for \emph{falsification-driven discovery} in which executable scientific artifacts and validity-constrained adversarial environments co-evolve under a fixed scientific protocol.
    \item We unify innovation, \emph{environment synthesis}, \emph{validity-constrained falsification}, and \emph{mechanistic feedback} into a single loop, turning failure from an opaque score signal into structured scientific correction.
    \item We embed a \emph{scientific contract} into the discovery process itself, making adaptive autonomous discovery auditable, replayable, and resistant to protocol gaming.
\end{itemize}

If recent autonomous scientist systems showed that AI can search vastly faster than human researchers, DASES asks the more consequential question:
\emph{can an automated scientist survive being falsified as rigorously as it can search?}
We argue that this is the threshold between benchmark-optimized automation and scientifically trustworthy discovery.
\section{DASES - Dynamic Adversarial Scientific Environment Synthesis and Mechanistic Co-Evolution}
\label{sec:method}

DASES begins from a precise claim: for sufficiently capable autonomous research systems, the dominant failure mode is no longer merely weak hypothesis search, but \emph{cognitive overfitting to a static evaluator}. A frozen benchmark is a frozen exam. Once that exam is fixed, a powerful search process can exploit benchmark-specific regularities, proxy-lab artifacts, or implementation loopholes without learning the mechanism the task was meant to reveal. DASES therefore changes the unit of scientific progress. A candidate is not retained because it scores well on a pre-specified testbed, but because repeated, validity-preserving attempts to break it fail. Automated discovery is thus recast as \emph{falsification-driven co-evolution}: the artifact under study and the environments capable of refuting it are optimized against one another under a fixed scientific protocol.

This shift requires five conditions to hold together: a single editable scientific locus, an explicit adversary over environments, validity-constrained falsification, mechanism-level feedback, and a scientific contract that prevents protocol gaming. DASES is the minimal framework we propose that satisfies these requirements jointly.

\subsection{Scientific Task as a Constrained Falsification Problem}
\label{sec:method:setup}

We consider research settings in which a candidate \emph{scientific artifact} can be instantiated, executed, and measured. The artifact may be any executable hypothesis, such as a loss function, optimizer, architecture component, controller, or program transformation. A DASES task is specified by
\begin{equation}
\mathcal{T}
=
\bigl(
\mathcal{H},
\Phi,
\mathcal{P},
\mathcal{I},
\mathcal{E},
\mathcal{C},
\mathcal{R}
\bigr),
\label{eq:task_dases}
\end{equation}
where $\mathcal{H}$ is the hypothesis space, $\Phi$ is a constrained edit interface, $\mathcal{P}$ is the immutable base protocol, $\mathcal{I}$ encodes the task design intent, $\mathcal{E}$ is a family of adversarially synthesizable environments, $\mathcal{C}$ is the set of validity constraints those environments must satisfy, and $\mathcal{R}$ is a pre-registered robustness contract defining survival.

Among these components, $\Phi$ is foundational. It creates a \emph{single editable scientific locus}: the only authorized place where the candidate may intervene on the system under study. This is not an implementation convenience, but the first safeguard of the method. Without such a locus, an autonomous agent could improve results by modifying the training schedule, evaluation code, preprocessing pipeline, or other protected parts of the protocol. With $\Phi$, every round isolates the intended scientific variable. Given a hypothesis $h_t \in \mathcal{H}$, the instantiated artifact is
\begin{equation}
a_t
=
\operatorname{Run}\!\bigl(\mathcal{P}, \Phi(h_t)\bigr),
\label{eq:artifact_dases}
\end{equation}
with $\mathcal{P}$ held fixed.

The objective of discovery is therefore not single-distribution performance, but robustness under the strongest admissible refutations currently available:
\begin{equation}
h^\star
=
\operatorname*{arg\,max}_{h \in \mathcal{H}}
\;
\inf_{e \in \mathcal{E}(\mathcal{C})}
\;
U\!\left(
\operatorname{Eval}\!\bigl(\operatorname{Run}(\mathcal{P}, \Phi(h)), e\bigr)
\right),
\label{eq:minimax_dases}
\end{equation}
where $U(\cdot)$ aggregates structured evaluation outputs into a scalar robustness utility. Equation~\eqref{eq:minimax_dases} makes the scientific target explicit: DASES does not seek the artifact that looks strongest under a frozen exam, but the one that remains credible under the hardest \emph{valid} attempts to show that its success is spurious.

\subsection{DASES as a Closed Scientific Game}
\label{sec:method:agents}

DASES operationalizes Eq.~\eqref{eq:minimax_dases} as a closed loop among three roles. The first is the \emph{Innovator}, which proposes a new hypothesis, applies the authorized edit, executes the immutable protocol, and returns both the artifact and its provenance:
\begin{equation}
(a_t, \mu_t)
=
\mathcal{G}\!\bigl(\mathcal{T}, \mathcal{B}_{t-1}\bigr).
\label{eq:generator_dases}
\end{equation}
The provenance $\mu_t$ records the scientific evidence of the round: code diffs, run metadata, logs, intermediate metrics, and grounded literature notes. The Innovator may be highly capable and tool-using, but it is never allowed to act outside $\Phi$.

The second role is the \emph{Abyss Falsifier}. It is not a fixed benchmark or passive reviewer. Its sole purpose is to break the current candidate in scientifically meaningful ways. Conditioned on the task design intent $\mathcal{I}$, the artifact $a_t$, its provenance $\mu_t$, and the prior falsification trace $\tau_{<t}$, it synthesizes a candidate-specific suite
\begin{equation}
S_t
=
\mathcal{D}\!\bigl(\mathcal{I}, a_t, \mu_t, \tau_{<t}\bigr)
\subseteq
\mathcal{E}(\mathcal{C}).
\label{eq:suite_dases}
\end{equation}
This conditioning is critical. The evaluator does not merely become broader over time; it becomes more diagnostically precise, targeting the shortcut mechanisms, brittle assumptions, and explanatory mismatches most plausible for the current artifact.

The artifact is then evaluated on the synthesized suite,
\begin{equation}
m_t
=
\left\{
\operatorname{Eval}(a_t, e)
\;:\;
e \in S_t
\right\},
\label{eq:metrics_dases}
\end{equation}
producing structured diagnostics rather than a single scalar score.

The third role is the \emph{Mechanistic Causal Extractor}. A system that can only report that a candidate failed, but cannot explain why, remains trapped in a stronger form of trial-and-error. DASES therefore requires each falsification trace to be converted into a mechanism-level account:
\begin{equation}
f_t
=
\mathcal{M}\!\bigl(m_t, S_t, \mu_t\bigr)
=
(\omega_t, \Gamma_t),
\label{eq:feedback_dases}
\end{equation}
where $\omega_t$ is the most plausible explanation of failure or survival, and $\Gamma_t$ is the minimal next-round correction expressed strictly in the language of the editable interface. $\mathcal{M}$ may be a separate module or partially fused with the Falsifier, but its function is irreducible: it turns evaluation into diagnosis, and diagnosis into a constrained scientific update. Scientific refinement in DASES is therefore not proposal plus scoring; it is proposal, targeted refutation, and mechanism-level correction inside a fixed protocol.

\subsection{Dynamic Adversarial Environment Synthesis}
\label{sec:method:env}

The central methodological move in DASES is to elevate the environment itself into a first-class scientific object. The Falsifier does not merely resample more data from a larger pool. It constructs executable counterexamples designed to answer a sharper question: has the artifact captured the intended mechanism, or has it learned a shortcut that merely happened to work under the original conditions? Each adversarial environment is represented as
\begin{equation}
e = (\pi, \theta, s, \nu),
\label{eq:envspec_dases}
\end{equation}
where $\pi$ is the generation program or protocol, $\theta$ its parameters, $s$ a random seed, and $\nu$ auxiliary validation metadata. This representation makes every falsification attempt replayable, auditable, and refinable across rounds.

Adversariality alone, however, is not enough. An environment is scientifically admissible only if it preserves the task semantics or constitutes a legitimate edge case under the intended problem definition. DASES therefore enforces
\begin{equation}
\operatorname{Valid}(e; \mathcal{I}, \mathcal{C}) = 1,
\label{eq:valid_dases}
\end{equation}
which states that every synthesized environment must satisfy the task-specific validity constraints. These constraints define what may be perturbed, what must be preserved, and what counts as a lawful stress condition. This is the decisive distinction from unconstrained adversarial attack generation: the objective is not to manufacture arbitrary failure, but to synthesize \emph{scientifically admissible} counterexamples.

For diagnostic coverage, the synthesized suite is structured rather than monolithic:
\begin{equation}
S_t
=
S_t^{\mathrm{inv}}
\;\cup\;
S_t^{\mathrm{cf}}
\;\cup\;
S_t^{\mathrm{tail}}
\;\cup\;
S_t^{\mathrm{comp}},
\label{eq:decomp_dases}
\end{equation}
where $S_t^{\mathrm{inv}}$ contains invariance tests, $S_t^{\mathrm{cf}}$ contains counterfactual interventions, $S_t^{\mathrm{tail}}$ contains edge and tail stressors, and $S_t^{\mathrm{comp}}$ contains compositional adversaries. The distinction is scientific rather than cosmetic. Failure under invariance tests indicates sensitivity to nuisance variation that should have been ignored. Failure under counterfactual intervention is stronger: it suggests that the artifact may be relying on the wrong signal altogether. Failure concentrated in tail regimes can instead indicate insufficient margin, poor calibration, or instability invisible in average-case evaluation. Compositional failures reveal interactions that single-axis stressors cannot expose. DASES uses this decomposition to turn evaluation into a map of causal and robustness liabilities rather than a mere ranking device.

The synthesis process is candidate-conditioned in a stronger sense than ordinary robustness testing. It jointly uses the task design intent $\mathcal{I}$, which specifies what should remain causally decisive; the artifact and its provenance, which reveal what the current solution is likely exploiting; and the historical trace, which records which failure modes have already been exposed and which remain unresolved. The environment frontier therefore does not merely become harder; it becomes progressively more precise.

\subsection{Mechanistic Co-Evolution and Survival Semantics}
\label{sec:method:coev}

DASES treats failure as structured evidence. After each round, the diagnosis $f_t=(\omega_t,\Gamma_t)$ is written into a bounded falsification memory,
\begin{equation}
\mathcal{B}_t
=
\operatorname{Update}\!\bigl(\mathcal{B}_{t-1}, f_t\bigr),
\label{eq:memory_dases}
\end{equation}
and the next hypothesis is generated conditionally on that memory:
\begin{equation}
h_{t+1}
\sim
p_{\mathcal{G}}\!\bigl(h \mid \mathcal{T}, \mathcal{B}_t\bigr).
\label{eq:next_dases}
\end{equation}
This is what makes DASES a co-evolution framework rather than a wrapper around harder testing. Over rounds, the system is not simply searching for higher scores; it is progressively eliminating explanatory stories that fail under admissible refutation. What remains are candidates that have survived increasingly targeted attempts to show that their success is non-mechanistic.

The acceptance criterion is therefore defined by survival, not by mean performance. Let the robustness contract be
\[
\mathcal{R}=\{(\rho_j,b_j)\}_{j=1}^J,
\]
where each $\rho_j$ measures a task-relevant violation and each $b_j$ specifies its admissible bound. The round verdict is
\begin{equation}
v_t
=
\mathbf{1}\!\left[
\forall j \in \{1,\dots,J\},
\;
\rho_j(m_t) \le b_j
\right].
\label{eq:verdict_dases}
\end{equation}
The functions $\rho_j$ may encode worst-slice error, invariance gap, counterfactual sensitivity, calibration under stress, tail-risk exposure, or other scientifically meaningful violations. A candidate is therefore accepted not because it is best on average, but because the strongest admissible attempts to invalidate its claim have, at the current frontier, failed.

\subsection{Scientific Contract: Integrity, Anti-Cheating, and Auditability}
\label{sec:method:integrity}

Adaptive evaluation raises epistemic pressure, but it also raises the stakes of protocol tampering. A sufficiently capable autonomous system may improve apparent performance by editing protected code, contaminating data, exploiting stale runtime state, or importing unsupported claims into the search trace. DASES addresses this by embedding a scientific contract directly into the method.

Let $\Omega$ denote the protected components of the base protocol and let $\Lambda_{\Phi}\subset\Omega$ denote the authorized editable locus. DASES enforces protocol immutability outside that locus through
\begin{equation}
\operatorname{hash}_t(x)
=
\operatorname{hash}_0(x),
\qquad
\forall x \in \Omega \setminus \Lambda_{\Phi},
\label{eq:immutability_dases}
\end{equation}
before and after major phases of each round. Any violation invalidates the round. This guarantee ensures that any apparent gain must originate from the scientific object under study rather than from silent goalpost movement.

The same contract also requires isolated execution where necessary so that crashes, stale states, and partial corruption cannot leak across rounds; replayable environment specifications so that every falsification attempt can be rerun exactly; and full provenance so that each round records the proposal, evidence, generated environments, metrics, verdict, diagnosis, and artifacts required for post-hoc analysis. When the Innovator relies on external literature or prior documents, DASES further stores grounded notes or verifiable excerpts instead of allowing unsupported claims to silently shape the future search state. The framework therefore returns not only a surviving candidate, but also a falsification trace explaining why weaker candidates did not deserve to survive.

\subsection{Round-Based Orchestration and Scope}
\label{sec:method:orchestration}

DASES proceeds for at most $T$ rounds. It begins by instantiating the task tuple in Eq.~\eqref{eq:task_dases}, recording integrity manifests for protected components, and initializing the falsification memory $\mathcal{B}_0$. At round $t$, the system first verifies the scientific contract. The Innovator then proposes $h_t$, applies the authorized edit, executes $\mathcal{P}$, and emits $(a_t,\mu_t)$. The contract is checked again to ensure that no protected component outside $\Lambda_{\Phi}$ has changed. The Abyss Falsifier synthesizes the candidate-conditioned suite $S_t$ and evaluates the artifact to obtain $m_t$. The Mechanistic Extractor converts the resulting trace into $(\omega_t,\Gamma_t)$, updates the falsification memory, and determines the verdict through Eq.~\eqref{eq:verdict_dases}. If $v_t=1$, the candidate is accepted as having survived the current adversarial frontier; otherwise the loop continues. The final output includes the surviving artifact, the complete falsification trace, all environment specifications, and the provenance required for reproduction and audit.

This orchestration also reveals the scope of the framework. A problem is DASES-compatible exactly when three conditions can be specified: an editable scientific locus $\Phi$ that isolates the object of intervention, an immutable protocol $\mathcal{P}$ that defines fair comparison, and a validity-constrained family of adversarial environments $\mathcal{E}(\mathcal{C})$ capable of producing meaningful refutations. Once these conditions exist, the problem can be cast as falsification-driven co-evolution. The deeper object of optimization is then not only a candidate artifact, but the pair consisting of a mechanistic claim and an increasingly capable process for proving that claim false. Scientific progress occurs when the second can no longer easily invalidate the first under admissible conditions.
\section{Experiments}
\label{sec:exp}

Our experiments are designed to test the central methodological claim of DASES: \emph{whether adaptive, validity-constrained falsification can discover a scientific artifact that static validation would have accepted too early, and whether the surviving artifact transfers beyond the discovery environment}. We instantiate DASES on a deliberately narrow but scientifically revealing problem: discovering a better and more general loss function, including regularization, for image classification. The narrowness is intentional rather than limiting. Throughout the entire discovery game, the backbone architecture, optimizer, augmentation policy, data pipeline, and training schedule are frozen. The \emph{only} editable scientific locus is the loss applied to the fixed penultimate representation and logits. This provides a direct realization of the edit interface $\Phi$ in Section~\ref{sec:method}: any performance difference must arise from the scientific object under study itself, not from hidden changes to architecture, optimization, or evaluation.

This point is important for interpreting the contribution correctly. The main claim of the experiment is not merely that an automated search procedure can stumble upon a stronger loss. It is that DASES can keep a candidate under progressively sharper, scientifically admissible falsification pressure until weaker explanatory stories are eliminated, and that the first artifact surviving this process is qualitatively different from what a static evaluator would have selected.

\subsection{DASES instantiation for loss discovery}
\label{sec:exp:instantiation}

We instantiate the task tuple in Eq.~\eqref{eq:task_dases} as follows. The hypothesis space $\mathcal{H}$ consists of executable loss functions with optional regularization terms. The edit interface $\Phi$ exposes only the loss definition. The immutable base protocol $\mathcal{P}$ fixes the classifier, optimizer, learning-rate schedule, training epochs, data transformations, random seeds, and evaluation logic. The design intent $\mathcal{I}$ is explicit: the classifier should predict the label from \emph{foreground shape geometry}, not from background statistics or incidental visual cues. The adversarial environment family $\mathcal{E}(\mathcal{C})$ consists of code-generated test distributions that preserve label-defining shape semantics while varying nuisance factors, edge conditions, and counterfactual background assignments. The robustness contract $\mathcal{R}$ is pre-registered before the discovery loop begins and includes four classes of violations: background counterfactual error, paired invariance inconsistency, worst-slice tail error, and compositional OOD error. 
% Exact metric definitions, slice construction rules, and admissible thresholds are specified in Appendix~X. 
A candidate survives a round only if all contract terms remain below their pre-registered bounds.

This setup directly realizes the three DASES roles. The \emph{Innovator} proposes a loss candidate, trains the fixed classifier under $\mathcal{P}$, and returns the artifact together with full provenance. The \emph{Abyss Falsifier} analyzes the logic of the training distribution, identifies scientifically lawful weaknesses, synthesizes candidate-conditioned test environments satisfying $\operatorname{Valid}(e;\mathcal{I},\mathcal{C})=1$, and evaluates the artifact on the resulting suites. The \emph{Mechanistic Causal Extractor} converts the falsification trace into an explanation of why the candidate failed or survived and outputs the minimal next-round correction expressed strictly in the language of the editable locus: the loss. The experiment is therefore not a loose narrative around search and testing; it is an operational realization of the closed scientific game in Section~\ref{sec:method:agents}--\ref{sec:method:coev}.

\subsection{A deceptive but lawful discovery environment}
\label{sec:exp:env}

The discovery environment is a synthetic four-class image-classification task in which the true causal rule is simple but the training distribution is intentionally deceptive. Labels $Y\in\{0,1,2,3\}$ are determined solely by foreground geometry: circle, square, triangle, and an ``other'' class consisting of polygons with at least five sides. This foreground geometry is the intended mechanism. However, the training and validation distributions inject a strong shortcut: each class is associated with a background color--texture family with high probability. These backgrounds are not trivial templates; within each family they remain highly variable through gradients, layered textures, speckle, low-frequency cloud structure, patch mixing, and related stochastic effects. Foreground shapes also vary in size, position, rotation, line width, filling, corner rounding, mild irregularity, and projection, with additional imaging perturbations such as noise, blur, contrast jitter, compression artifacts, and vignetting. Train and validation share the same correlation mechanism.

The purpose of this environment is not to mimic the pixel distribution of real benchmarks. Its function is more precise: to isolate, in a controlled and auditable way, a class of failure modes that also arises in real learning systems---shortcut reliance, shallow geometric proxies, and collapse under compositional nuisance interactions. The environment is therefore a \emph{falsification laboratory}, not a surrogate benchmark. Static validation is expected to be misleading here for exactly the reason argued in Section~\ref{sec:intro}: the shortcut is cheaper than the intended mechanism, so a sufficiently capable searcher can look scientifically successful without learning the right thing.

The Falsifier is not allowed to generate arbitrary attacks. Every synthesized environment must preserve the rule that label is determined by foreground shape, or else instantiate a legitimate edge case under that rule. This is the admissibility condition in Eq.~\eqref{eq:valid_dases}. The objective is not to break the classifier by any means available, but to separate genuine mechanism learning from success that is contingent on artifacts of the original training distribution.

\subsection{Falsification-driven co-evolution in the discovery lab}
\label{sec:exp:trajectory}

The first falsification frontier targets the most obvious weakness of the training distribution: the induced correlation between label and background family. The Falsifier therefore begins with a counterfactual suite $S^{\mathrm{cf}}$ that preserves foreground shape while reversing, swapping, or neutralizing the background assignment. A circle may be rendered against a background family most commonly associated with a square, or against a neutral background that removes the original correlation altogether. These are lawful counterexamples because the label-defining geometry is unchanged. Early loss candidates fail sharply on this suite, showing that they have learned background statistics rather than the intended geometric mechanism.

Under an ordinary workflow, a strong train/validation result would already look convincing. Under DASES, it is not enough. After several failures and minimal corrections, the Innovator converges to the strongest standard baseline under the frozen protocol: plain cross-entropy (CE). CE improves substantially on the first counterfactual frontier. This is already a meaningful finding: DASES is not merely destructive. It rejects weaker candidates and forces the search toward stronger ones. But the system does not stop when the first failure mode is repaired.

The second falsification frontier arises from a subtler weakness in the original environment. Once background correlation is neutralized, the training generator still tends to present the foreground object as visually \emph{salient}: typically single-object, reasonably isolated, and often distinguishable by coarse silhouette, contrast, or occupancy cues. A model can therefore survive the first frontier without learning shape geometry in a stable mechanistic sense. It may instead rely on shallow proxies such as rough interior mass, easy boundary contrast, or simple edge-density statistics. These cues are closer to the true mechanism than background color, but they still do not constitute the mechanism the task is meant to reward.

The Falsifier addresses this by synthesizing an invariance-heavy second suite $S^{\mathrm{inv}}$ that removes those easy proxies while preserving class-defining shape. It generates examples with matched global color statistics, reduced foreground--background contrast, fill inversion, outline-only renderings, extreme but valid stroke widths, area-matched cross-class comparisons, and placements that push the object toward the image boundary without destroying recognizability. CE now drops substantially. The Mechanistic Extractor returns a sharper diagnosis: the model is no longer primarily shortcutting through background identity, but its representation of shape remains geometrically brittle. It is solving the task by a shallow notion of visual resemblance rather than by a stable encoding of class-defining geometry.

The Innovator then moves to the next natural candidate, cross-entropy with L2 regularization. This recovers some robustness and improves stability, but again reaches a ceiling. The third falsification frontier is where the DASES mechanism becomes decisive. The Falsifier revisits the training generator and identifies a weakness not yet exhausted by the first two suites: the nuisance factors in training are broad but mostly \emph{factorized} and of moderate strength. The model sees many valid perturbations, but often one at a time or in relatively benign combinations. A classifier can therefore survive isolated interventions while remaining fragile at the intersections where several lawful difficulties co-occur.

To expose this, the Falsifier expands the testbed with a compositional tail suite $S^{\mathrm{tail}}\cup S^{\mathrm{comp}}$. In this suite, multiple valid stressors are composed in the same example: background reassignment, low contrast, thin or partially broken outlines, off-center or boundary-touching placement, compression or blur, and local clutter or partial occlusion, all under the constraint that the foreground geometry remains label-preserving and recoverable by a human observer. This is not arbitrary attack generation. It is a lawful probe of whether the learned mechanism survives exactly where shortcuts and brittle proxies tend to fail. Under this frontier, CE$+$L2 breaks down again, especially on worst-slice error and paired consistency. The error pattern is not random: confidence remains high while evidence becomes structurally weak, margins are uneven across classes, and nuisance interactions collapse the separation that looked adequate under simpler tests.

\subsection{The artifact selected by the falsification trace}
\label{sec:exp:fngce}

The final candidate is therefore not introduced as another generic regularizer. It is forced by the accumulated diagnostic trace. The Extractor identifies two unresolved liabilities left by CE$+$L2. First, CE can still increase confidence through feature-norm inflation, allowing the classifier to improve scores by ``winning through length'' rather than by stable directional geometry. Second, the learned representation remains anisotropic and partially redundant, so class separation degrades when counterfactual and compositional stressors interact. The minimal correction consistent with the editable locus is thus:
\begin{equation}
\mathcal{L}_{\mathrm{FNG\text{-}CE}}
=
\mathcal{L}_{\mathrm{CE}}
+
\lambda_1 \frac{1}{N}\sum_{i=1}^{N}\left(\|f_i\|_2-c\right)^2
+
\lambda_2
\left\|
\frac{1}{N}\sum_{i=1}^{N}(f_i-\mu)(f_i-\mu)^\top-I
\right\|_F^2
+
\lambda_3\|W\|_2^2,
\label{eq:fngce_exp_final}
\end{equation}
where $f_i$ denotes the penultimate feature, $\mu=\frac{1}{N}\sum_i f_i$, and $W$ denotes the classifier weights. We call the resulting artifact \textbf{FNG-CE} (\emph{Feature Norm and Geometry Regularized Cross-Entropy}).

A likely reviewer concern is that the individual ingredients are not all unprecedented in isolation. That is not the claim of the paper. The claim is methodological and stronger: under a fixed protocol and an adaptive falsification game, DASES identifies this \emph{particular minimal combination} as the first artifact that satisfies the pre-registered robustness contract. To make this point auditable rather than rhetorical, we include both component ablations and comparisons against strong regularized and geometry-aware baselines in Table~X and Appendix~\ref{sec:exp_setup_app}, including CE, CE$+$L2, label smoothing, focal-style variants, center-style regularization, margin-based objectives, and other representative alternatives. The ablations show that neither norm control nor covariance regularization alone is sufficient to survive the full falsification frontier; the full combination is required to close the remaining failure gap.

Each term in Eq.~\eqref{eq:fngce_exp_final} directly answers a failure mode exposed by the Falsifier. The norm penalty suppresses confidence inflation by constraining features toward a stable radius, forcing discrimination to rely more on direction than arbitrary magnitude. The covariance regularizer discourages collapsed or highly redundant feature geometry, encouraging a more uniform and information-rich representation that better survives invariance and compositional stress. The L2 term remains useful as a standard capacity-control mechanism, but DASES shows that capacity control alone is not enough. FNG-CE is therefore not a decorative post-hoc design. It is the mechanistically motivated endpoint of the co-evolutionary process.

\subsection{Results in the discovery lab}
\label{sec:exp:toyresults}

\begin{figure}[!t]
    \centering
    \includegraphics[width=\linewidth]{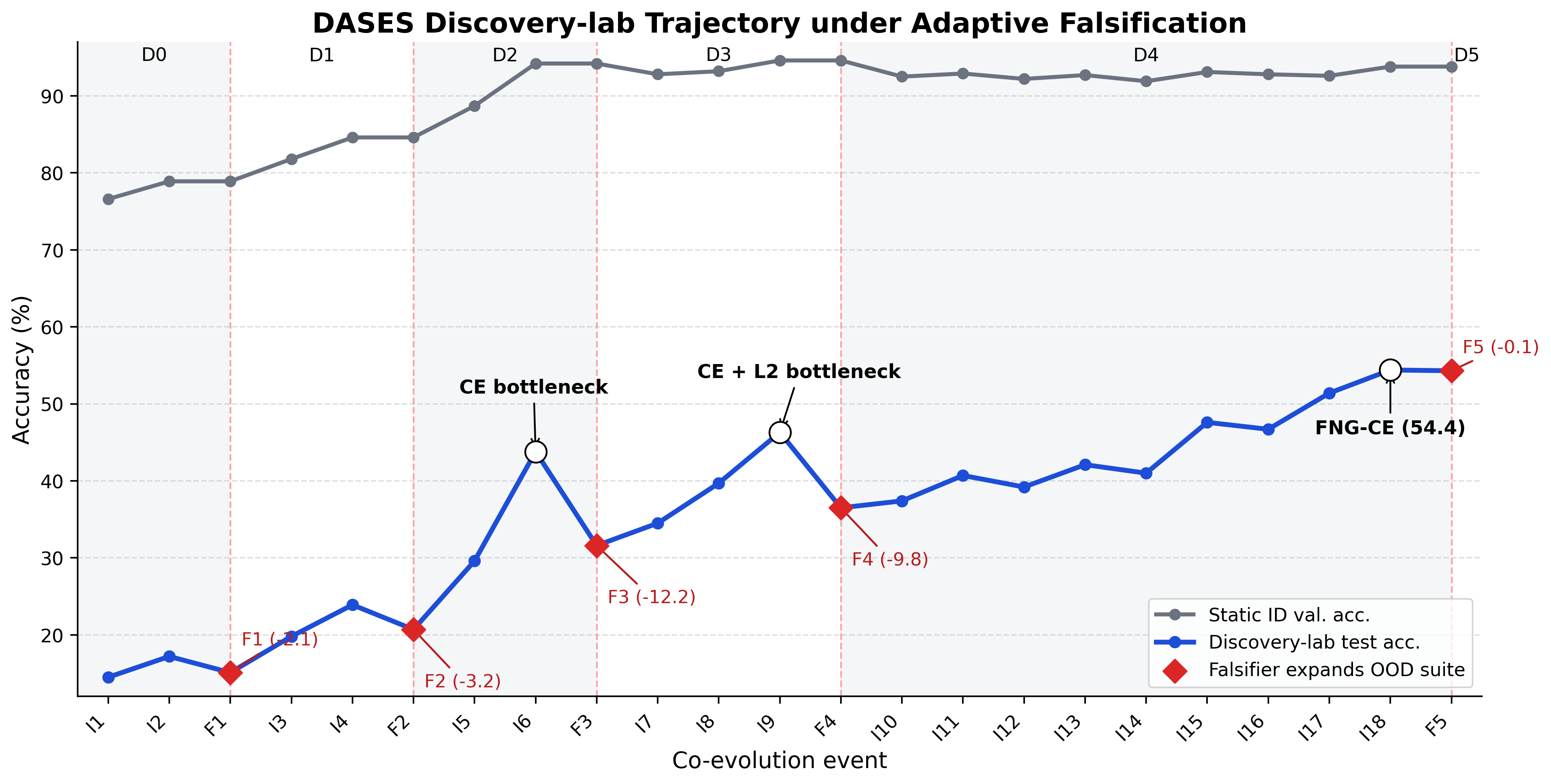}
    \vspace{-5mm}
    \caption{DASES co-evolution trajectory in the discovery lab. The static ID validation accuracy remains comparatively high throughout the search, but each falsifier expansion of the lawful OOD frontier induces a sharp test-score drop that exposes a new hidden failure mode. CE first mitigates background shortcut reliance, CE$+$L2 further improves robustness under geometry-focused stress, and FNG-CE is the first candidate to reach the highest plateau while remaining essentially stable under the final falsifier expansion.}
    \label{fig:discovery_lab_trajectory}
\end{figure}

Table \ref{tab:dases_discovery_trajectory} and Fig.~\ref{fig:discovery_lab_trajectory} report the full stage-wise results in the discovery environment. The main pattern is not merely that FNG-CE obtains the highest final score. It is that each newly synthesized suite reveals a distinct failure mode that static validation would not have exposed, and that the surviving candidate aligns with those failures in exactly the order predicted by DASES. Early candidates fail under background counterfactuals. CE largely repairs that weakness but fails under geometry-focused invariance stress. CE$+$L2 partially repairs the second weakness but remains fragile under compositional tail interactions. FNG-CE is the first candidate, among the losses we consider under the fixed training budget and protocol, to satisfy the full robustness contract across all currently synthesized suites. Its gains are visible not only in aggregate test accuracy, but also in lower counterfactual error, stronger paired consistency, improved worst-slice robustness, and reduced compositional OOD failure.

% \vspace{-1em}
\begin{table}[htbp]
\centering
\scriptsize
\setlength{\tabcolsep}{4.8pt}
\renewcommand{\arraystretch}{1.10}
\begin{threeparttable}
\caption{Predicted stage-wise co-evolution trajectory in the DASES discovery lab. Each falsifier step expands the lawful OOD frontier, so the associated test drop reflects stricter admissible evaluation rather than retraining instability.}
\label{tab:dases_discovery_trajectory}
\begin{tabular}{cclcccc}
\toprule
Code & Role & Candidate / action & Frontier & Static ID Val. (\%) & Test (\%) & $\Delta$ Test \\
\midrule
I1  & Innovator & Naive custom loss A                              & $D_0$ & 76.6 & 14.5 & -- \\
I2  & Innovator & Naive custom loss B                              & $D_0$ & 78.9 & 17.2 & +2.7 \\
\rowcolor{gray!10}
F1  & Falsifier & Add neutral-background counterfactuals           & $D_1$ & 78.9 & 15.1 & -2.1 \\
I3  & Innovator & Naive custom loss C                              & $D_1$ & 81.8 & 19.8 & +4.7 \\
I4  & Innovator & Shortcut-penalty heuristic                       & $D_1$ & 84.6 & 23.9 & +4.1 \\
\rowcolor{gray!10}
F2  & Falsifier & Add harder background-family swaps               & $D_2$ & 84.6 & 20.7 & -3.2 \\
I5  & Innovator & Weak de-correlation loss                         & $D_2$ & 88.7 & 29.6 & +8.9 \\
I6  & Innovator & \textbf{CE} (first bottleneck)                   & $D_2$ & \textbf{94.2} & \textbf{43.8} & +14.2 \\
\rowcolor{gray!10}
F3  & Falsifier & Add invariance-heavy geometry stress             & $D_3$ & 94.2 & 31.6 & -12.2 \\
I7  & Innovator & CE + label smoothing                             & $D_3$ & 92.8 & 34.5 & +2.9 \\
I8  & Innovator & CE + center penalty                              & $D_3$ & 93.2 & 39.7 & +5.2 \\
I9  & Innovator & \textbf{CE + L2} (second bottleneck)             & $D_3$ & \textbf{94.6} & \textbf{46.3} & +6.6 \\
\rowcolor{gray!10}
F4  & Falsifier & Add compositional tail interactions              & $D_4$ & 94.6 & 36.5 & -9.8 \\
I10 & Innovator & CE + focal reweighting                           & $D_4$ & 92.5 & 37.4 & +0.9 \\
I11 & Innovator & CE + label smoothing + L2                        & $D_4$ & 92.9 & 40.7 & +3.3 \\
I12 & Innovator & CE + margin term                                 & $D_4$ & 92.2 & 39.2 & -1.5 \\
I13 & Innovator & CE + center-style regularization                 & $D_4$ & 92.7 & 42.1 & +2.9 \\
I14 & Innovator & CE + angular margin                              & $D_4$ & 91.9 & 41.0 & -1.1 \\
I15 & Innovator & CE + feature-norm control                        & $D_4$ & 93.1 & 47.6 & +6.6 \\
I16 & Innovator & CE + covariance shaping                          & $D_4$ & 92.8 & 46.7 & -0.9 \\
I17 & Innovator & Norm + covariance (no L2)                        & $D_4$ & 92.6 & 51.4 & +4.7 \\
I18 & Innovator & \textbf{FNG-CE}                                  & $D_4$ & \textbf{93.8} & \textbf{54.4} & +3.0 \\
\rowcolor{gray!10}
F5  & Falsifier & Add residual lawful edge cases; no new breach   & $D_5$ & 93.8 & \textbf{54.3} & -0.1 \\
\bottomrule
\end{tabular}
\begin{tablenotes}[flushleft]
\footnotesize
\item $D_0$: initial shortcut-biased hold-out;
$D_1$: $D_0$ + neutral-background counterfactuals;
$D_2$: $D_1$ + harder background-family swaps;
$D_3$: $D_2$ + invariance-heavy geometry stressors;
$D_4$: $D_3$ + compositional/tail stressors;
$D_5$: $D_4$ + residual lawful edge cases.
\end{tablenotes}
\end{threeparttable}
\vspace{-5mm}
\end{table}

Just as importantly, the process terminates for the right reason. After additional rounds, the Falsifier is unable, \emph{within the admissible synthesis space induced by $\mathcal{I}$ and $\mathcal{C}$, the current falsifier search budget, and the pre-registered robustness contract}, to generate a new suite that opens another material violation. Newly proposed environments either reinstantiate already-resolved weaknesses or produce only marginal degradation without breaching the contract thresholds. We do \emph{not} interpret this as a universal proof that no stronger falsifier could ever exist. The claim is narrower and scientifically appropriate: within the present task definition and lawful synthesis space, the currently accessible falsification frontier has been exhausted to the point that the retained artifact survives all material admissible refutations we can construct. Under DASES semantics, this is precisely what acceptance is meant to mean.

\subsection{Transfer beyond the discovery environment}
\label{sec:exp:transfer}

A decisive test is whether the retained artifact is merely adapted to the synthetic falsification laboratory that produced it. We therefore transfer FNG-CE \emph{without changing its analytic form} to standard natural-image classification benchmarks, using standard ResNet-18 and ResNet-50 backbones, with full architecture and optimization details deferred to Appendix~X\ref{sec:exp_setup_app}. The evaluation splits are exactly those reported in Tables~\ref{LE_result_r18} and~\ref{LE_result_r50}: ResNet-18 on CIFAR10, CIFAR100, DTD, CUBirds, VGGFlower, and TrafficSigns; and ResNet-50 on ImageNet, CIFAR10, CIFAR100, DTD, CUBirds, VGGFlower, and TrafficSigns.
\vspace{-1mm}
\begin{table}[H]
\centering
\small
\setlength{\tabcolsep}{5.5pt}
\renewcommand{\arraystretch}{1.05}
\resizebox{0.92\linewidth}{!}{%
\begin{tabular}{l|cccccc}
\toprule
Methods & CIFAR10 & CIFAR100 & DTD & CUBirds & VGGFlower & TrafficSigns \\
\midrule
CE            & \second{86.36} & \second{59.72} & 14.79           & \second{13.57}           & \second{68.85} & \second{94.73} \\
CE+L2         & 86.23          & 57.95          & \second{15.53} & 13.45 & 68.01          & 94.25 \\
FNG-CE (ours) & \best{86.83}   & \best{60.04}   & \best{18.51}   & \best{13.86}   & \best{69.17}   & \best{94.84} \\
\midrule
Gain vs.\ CE (pp) & +0.47 & +0.32 & +3.72 & +0.29 & +0.32 & +0.11 \\
\bottomrule
\end{tabular}%
}
\caption{Evaluation accuracy (\%) on six datasets using the standard ResNet-18 backbone. FNG-CE achieves the best result on every dataset. Bold indicates the best result in each column, underlining indicates the runner-up, and the last row reports the absolute accuracy gain of FNG-CE over the CE baseline in percentage points (pp).}
\label{LE_result_r18}
\end{table}

\vspace{-2mm}

\begin{table}[H]
\centering
\small
\setlength{\tabcolsep}{5.2pt}
\renewcommand{\arraystretch}{1.05}
\resizebox{0.98\linewidth}{!}{%
\begin{tabular}{l|ccccccc}
\toprule
Methods & ImageNet & CIFAR10 & CIFAR100 & DTD & CUBirds & VGGFlower & TrafficSigns \\
\midrule
CE            & \second{70.73} & \second{87.42} & \second{61.28} & \second{18.46} & \second{13.98} & \second{64.37} & \second{94.71} \\
CE+L2         & 68.24          & 86.24          & 59.59          & 16.65          & 13.58          & 62.46          & 93.88 \\
FNG-CE (ours) & \best{71.56}   & \best{87.94}   & \best{61.75}   & \best{19.10}   & \best{14.11}   & \best{65.05}   & \best{94.94} \\
\midrule
Gain vs.\ CE (pp) & +0.83 & +0.52 & +0.47 & +0.64 & +0.13 & +0.68 & +0.23 \\
\bottomrule
\end{tabular}%
}
\caption{Evaluation accuracy (\%) on seven datasets using the standard ResNet-50 backbone. FNG-CE achieves the best result on every dataset. Bold indicates the best result in each column, underlining indicates the runner-up, and the last row reports the absolute accuracy gain of FNG-CE over the CE baseline in percentage points (pp).}
\label{LE_result_r50}
\end{table}

\vspace{-1mm}

The transfer pattern is not marginal or selective; it is uniform. FNG-CE achieves the best result on every dataset under both backbones. Relative to CE, the ResNet-18 gains are +0.47 pp on CIFAR10, +0.32 on CIFAR100, +3.72 on DTD, +0.29 on CUBirds, +0.32 on VGGFlower, and +0.11 on TrafficSigns. Under ResNet-50, the gains remain positive across the board: +0.83 pp on ImageNet, +0.52 on CIFAR10, +0.47 on CIFAR100, +0.64 on DTD, +0.13 on CUBirds, +0.68 on VGGFlower, and +0.23 on TrafficSigns. Just as revealing, CE+L2 does \emph{not} exhibit this pattern: it trails CE on five of six ResNet-18 transfers and on all seven ResNet-50 transfers, whereas FNG-CE remains consistently superior.

This is the strongest possible outcome for the argument of this paper. The discovery environment was never meant to approximate the full natural-image statistics of these benchmarks. Its purpose was narrower and more scientific: to isolate, under auditable control, whether a candidate had actually become less dependent on shortcuts, less geometrically brittle, and more stable under lawful nuisance composition. The fact that the artifact selected by that falsification trace continues to improve broad real-image benchmarks indicates that DASES is not merely engineering a better synthetic test score. It is surfacing a more transferable loss-level inductive bias.

This closes the experimental loop. Static validation would have favored an earlier and weaker explanation of success. DASES instead keeps the candidate under escalating admissible refutation until the remaining survivor is not only harder to break inside the lab, but also stronger outside it. That is exactly the distinction the method is designed to enforce: not benchmark success, but mechanism that continues to matter after the discovery environment is left behind.

% \paragraph{Supervised training.}
% We perform standard supervised learning on ImageNet-1K~\cite{he2016deep}, CIFAR10~\cite{krizhevsky2009learning}, CIFAR100~\cite{coates2011analysis}, DTD~\cite{cimpoi2014describing}, CUBirds~\cite{wah2011caltech}, VGGFlower~\cite{nilsback2008automated}, and TrafficSigns~\cite{houben2013detection} using ResNet18 and ResNet50 backbones. All models are trained for 50 epochs with batch size 256, Adam optimizer, and learning rate \(3\times10^{-4}\). We evaluate two objectives: plain cross-entropy (CE) and CE with L2 regularization.

% \paragraph{Evaluation protocol.}
% For each dataset and backbone, we report the test accuracy at the final epoch. Only the last checkpoint is saved. In total, we run \(6\times2\times2=24\) experiments (6 datasets, 2 backbones, 2 losses: CE and CE+L2), and aggregate all results into a unified summary table.

\section{Conclusion}

We argued that the core bottleneck of autonomous scientific discovery is no longer search alone, but the static evaluator it searches against. Once evaluation is frozen, a capable system can optimize to the exam faster than it uncovers the intended mechanism. DASES changes that contract. It defines progress not as performance on a fixed testbed, but as survival under adaptive, validity-constrained falsification. The retained artifact is therefore not the one that merely scores highest; it is the first one that remains credible after repeated lawful attempts to break it.

Our experiments make this methodological claim concrete. In the discovery lab, each falsifier expansion exposes a hidden failure mode that static validation would have missed, and the final artifact is selected by the falsification trace rather than by headline in-distribution accuracy. The resulting loss, FNG-CE, then transfers beyond the synthetic lab and consistently outperforms CE and CE+L2 across all controlled benchmark comparisons we report.

DASES does not prove that no stronger falsifier could exist. Its claim is sharper and more scientific: autonomous discovery becomes more trustworthy when acceptance is earned by surviving the strongest admissible refutations we can currently construct, not by excelling on a frozen exam.

\bibliographystyle{plainnat}
\bibliography{references}

@article{hu2024adas,
  title={Automated design of agentic systems},
  author={Hu, Shengran and Lu, Cong and Clune, Jeff},
  journal={arXiv preprint arXiv:2408.08435},
  year={2024}
}

@inproceedings{wang2025evoagentx,
    title = "{E}vo{A}gent{X}: An Automated Framework for Evolving Agentic Workflows",
    author = "Wang, Yingxu  and
      Liu, Siwei  and
      Fang, Jinyuan  and
      Meng, Zaiqiao",
    editor = {Habernal, Ivan  and
      Schulam, Peter  and
      Tiedemann, J{\"o}rg},
    booktitle = "Proceedings of the 2025 Conference on Empirical Methods in Natural Language Processing: System Demonstrations",
    month = nov,
    year = "2025",
    address = "Suzhou, China",
    publisher = "Association for Computational Linguistics",
    url = "https://aclanthology.org/2025.emnlp-demos.47/",
    doi = "10.18653/v1/2025.emnlp-demos.47",
    pages = "643--655",
    ISBN = "979-8-89176-334-0"
}

@article{lu2024aiscientist,
  title={The ai scientist: Towards fully automated open-ended scientific discovery},
  author={Lu, Chris and Lu, Cong and Lange, Robert Tjarko and Foerster, Jakob and Clune, Jeff and Ha, David},
  journal={arXiv preprint arXiv:2408.06292},
  year={2024}
}

@article{yamada2025aiscientistv2,
  title={The ai scientist-v2: Workshop-level automated scientific discovery via agentic tree search},
  author={Yamada, Yutaro and Lange, Robert Tjarko and Lu, Cong and Hu, Shengran and Lu, Chris and Foerster, Jakob and Clune, Jeff and Ha, David},
  journal={arXiv preprint arXiv:2504.08066},
  year={2025}
}

@article{weng2025deepscientist,
  title={Deepscientist: Advancing frontier-pushing scientific findings progressively},
  author={Weng, Yixuan and Zhu, Minjun and Xie, Qiujie and Sun, Qiyao and Lin, Zhen and Liu, Sifan and Zhang, Yue},
  journal={arXiv preprint arXiv:2509.26603},
  year={2025}
}

@article{yu2025alpharesearch,
  title={Alpharesearch: Accelerating new algorithm discovery with language models},
  author={Yu, Zhaojian and Feng, Kaiyue and Zhao, Yilun and He, Shilin and Zhang, Xiao-Ping and Cohan, Arman},
  journal={arXiv preprint arXiv:2511.08522},
  year={2025}
}

@article{pu2026pievo,
  title={Principle-Evolvable Scientific Discovery via Uncertainty Minimization},
  author={Pu, Yingming and Lin, Tao and Chen, Hongyu},
  journal={arXiv preprint arXiv:2602.06448},
  year={2026}
}

@article{vitvitskyi2026mining,
  title={Mining Generalizable Activation Functions},
  author={Vitvitskyi, Alex and Boratko, Michael and Grcic, Matej and Pascanu, Razvan and Shah, Deep and Veli{\v{c}}kovi{\'c}, Petar},
  journal={arXiv preprint arXiv:2602.05688},
  year={2026}
}

@article{sargsyan2025structural,
  title={Structural Enforcement of Statistical Rigor in AI-Driven Discovery: A Functional Architecture},
  author={Sargsyan, Karen},
  journal={arXiv preprint arXiv:2511.06701},
  year={2025}
}

@inproceedings{he2016deep,
  title={Deep residual learning for image recognition},
  author={He, Kaiming and Zhang, Xiangyu and Ren, Shaoqing and Sun, Jian},
  booktitle={Proceedings of the IEEE conference on computer vision and pattern recognition},
  pages={770--778},
  year={2016}
}

@article{krizhevsky2009learning,
  title={Learning multiple layers of features from tiny images},
  author={Krizhevsky, Alex and Hinton, Geoffrey and others},
  journal={Master's thesis, University of Tront},
  year={2009},
  publisher={Citeseer}
}

@inproceedings{coates2011analysis,
  title={An analysis of single-layer networks in unsupervised feature learning},
  author={Coates, Adam and Ng, Andrew and Lee, Honglak},
  booktitle={Proceedings of the fourteenth international conference on artificial intelligence and statistics},
  pages={215--223},
  year={2011},
  organization={JMLR Workshop and Conference Proceedings}
}

@inproceedings{cimpoi2014describing,
  title={Describing textures in the wild},
  author={Cimpoi, Mircea and Maji, Subhransu and Kokkinos, Iasonas and Mohamed, Sammy and Vedaldi, Andrea},
  booktitle={Proceedings of the IEEE Conference on Computer Vision and Pattern Recognition},
  pages={3606--3613},
  year={2014}
}

@article{wah2011caltech,
  title={The caltech-ucsd birds-200-2011 dataset},
  author={Wah, Catherine and Branson, Steve and Welinder, Peter and Perona, Pietro and Belongie, Serge},
  year={2011},
  publisher={California Institute of Technology}
}

@inproceedings{nilsback2008automated,
  title={Automated flower classification over a large number of classes},
  author={Nilsback, Maria-Elena and Zisserman, Andrew},
  booktitle={2008 Sixth Indian Conference on Computer Vision, Graphics \& Image Processing},
  pages={722--729},
  year={2008},
  organization={IEEE}
}

@inproceedings{houben2013detection,
  title={Detection of traffic signs in real-world images: The German Traffic Sign Detection Benchmark},
  author={Houben, Sebastian and Stallkamp, Johannes and Salmen, Jan and Schlipsing, Marc and Igel, Christian},
  booktitle={The 2013 international joint conference on neural networks (IJCNN)},
  pages={1--8},
  year={2013},
  organization={Ieee}
}

@article{kingma2014adam,
  title={Adam: A method for stochastic optimization},
  author={Kingma, Diederik P and Ba, Jimmy},
  journal={arXiv preprint arXiv:1412.6980},
  year={2014}
}

@article{yang2025qwen3,
  title={Qwen3 technical report},
  author={Yang, An and Li, Anfeng and Yang, Baosong and Zhang, Beichen and Hui, Binyuan and Zheng, Bo and Yu, Bowen and Gao, Chang and Huang, Chengen and Lv, Chenxu and others},
  journal={arXiv preprint arXiv:2505.09388},
  year={2025}
}

@article{team2025gemma3,
  title={Gemma 3 technical report},
  author={Team, Gemma and Kamath, Aishwarya and Ferret, Johan and Pathak, Shreya and Vieillard, Nino and Merhej, Ramona and Perrin, Sarah and Matejovicova, Tatiana and Ram{\'e}, Alexandre and Rivi{\`e}re, Morgane and others},
  journal={arXiv preprint arXiv:2503.19786},
  year={2025}
}

@article{llama3.2,
  author       = {Meta},
  title        = {Llama 3.2: Revolutionizing edge AI and vision with open, customizable models},
  year         = {2024},
  url          = {https://ai.meta.com/blog/llama-3-2-connect-2024-vision-edge-mobile-devices/},
}

@article{singh2025openai,
  title={Openai gpt-5 system card},
  author={Singh, Aaditya and Fry, Adam and Perelman, Adam and Tart, Adam and Ganesh, Adi and El-Kishky, Ahmed and McLaughlin, Aidan and Low, Aiden and Ostrow, AJ and Ananthram, Akhila and others},
  journal={arXiv preprint arXiv:2601.03267},
  year={2025}
}

@inproceedings{pan2024plum,
  title={Plum: Prompt Learning using Metaheuristics},
  author={Pan, Rui and Xing, Shuo and Diao, Shizhe and Sun, Wenhe and Liu, Xiang and Shum, Kashun and Zhang, Jipeng and Pi, Renjie and Zhang, Tong},
  booktitle={Findings of the Association for Computational Linguistics: ACL 2024},
  pages={2177--2197},
  year={2024}
}

@inproceedings{xing2025re,
  title={Re-Align: Aligning vision language models via retrieval-augmented direct preference optimization},
  author={Xing, Shuo and Li, Peiran and Wang, Yuping and Bai, Ruizheng and Wang, Yueqi and Hu, Chan-Wei and Qian, Chengxuan and Yao, Huaxiu and Tu, Zhengzhong},
  booktitle={Proceedings of the 2025 Conference on Empirical Methods in Natural Language Processing},
  pages={2379--2397},
  year={2025}
}

@article{li2025safeflow,
  title={Safeflow: A principled protocol for trustworthy and transactional autonomous agent systems},
  author={Li, Peiran and Zou, Xinkai and Wu, Zhuohang and Li, Ruifeng and Xing, Shuo and Zheng, Hanwen and Hu, Zhikai and Wang, Yuping and Li, Haoxi and Yuan, Qin and others},
  journal={arXiv preprint arXiv:2506.07564},
  year={2025}
}

@article{li2026bibagent,
  title={BibAgent: An Agentic Framework for Traceable Miscitation Detection in Scientific Literature},
  author={Li, Peiran and Lin, Fangzhou and Xing, Shuo and Zheng, Xiang and Hong, Xi and Yang, Siyuan and Sun, Jiashuo and Tu, Zhengzhong and Ni, Chaoqun},
  journal={arXiv preprint arXiv:2601.16993},
  year={2026}
}

@article{li2026traversal,
  title={Traversal-as-Policy: Log-Distilled Gated Behavior Trees as Externalized, Verifiable Policies for Safe, Robust, and Efficient Agents},
  author={Li, Peiran and Sun, Jiashuo and Lin, Fangzhou and Xing, Shuo and Fu, Tianfu and Feng, Suofei and Ni, Chaoqun and Tu, Zhengzhong},
  journal={arXiv preprint arXiv:2603.05517},
  year={2026}
}

@article{dai2025language,
  title={A language anchor-guided method for robust noisy domain generalization},
  author={Dai, Zilin and Wang, Lehong and Lin, Fangzhou and Wang, Yidong and Li, Zhigang and Yamada, Kazunori D and Zhang, Ziming and Lu, Wang},
  journal={arXiv preprint arXiv:2503.17211},
  year={2025}
}

@article{lin2026position,
  title={Position: Human-Centric AI Requires a Minimum Viable Level of Human Understanding},
  author={Lin, Fangzhou and Ge, Qianwen and Xu, Lingyu and Li, Peiran and Gao, Xiangbo and Xing, Shuo and Yamada, Kazunori and Zhang, Ziming and Zhang, Haichong and Tu, Zhengzhong},
  journal={arXiv preprint arXiv:2602.00854},
  year={2026}
}

%%%%%%%%%%%%%%%%%%%%%%%%%%%%%%%%%%%%%%%%%%%%%%%%%%%%%%%%%%%%
\newpage
\appendix

% \section{Technical Appendices and Supplementary Material}
% Technical appendices with additional results, figures, graphs and proofs may be submitted with the paper submission before the full submission deadline (see above), or as a separate PDF in the ZIP file below before the supplementary material deadline. There is no page limit for the technical appendices.

\section{Experimental Setup}
\label{sec:exp_setup_app}

\subsection{Implementation and Environment}
We performed the same replacement for all the comparative losses in our experiments. We trained all these networks from scratch using PyTorch, such as learning rates, batch sizes and other factors for training networks were kept consistent with the same settings for fair comparisons: all networks are trained for 50 epochs with Adam optimizer~\cite{kingma2014adam}, learning rate \(3\times10^{-4}\), and batch size 256. No learning-rate scheduler or early stopping is used. We evaluate on the test split at every epoch and report the final-epoch test accuracy. Unless otherwise noted, we use a fixed random seed of 42 for data splitting and training reproducibility.

\subsection{Datasets}
We evaluate on six standard classification benchmarks:
ImageNet-1K~\cite{he2016deep}, CIFAR10~\cite{krizhevsky2009learning}, CIFAR100~\cite{coates2011analysis}, DTD~\cite{cimpoi2014describing}, CUBirds~\cite{wah2011caltech}, VGGFlower~\cite{nilsback2008automated}, and TrafficSigns~\cite{houben2013detection}.
The number of classes and split protocol are summarized in Table~\ref{tab:datasets_setup}.

\begin{table}[h]
\centering
\caption{Dataset setup used in this work.}
\label{tab:datasets_setup}
\begin{tabular}{lcc}
\hline
Dataset & \#Classes & Train/Test split protocol \\
\hline
CIFAR-10 & 10 & Official train/test split \\
CIFAR-100 & 100 & Official train/test split \\
DTD & 47 & Official split: \texttt{train1+val1} for training, \texttt{test1} for testing \\
CUB-200-2011 & 200 & Official split from \texttt{train\_test\_split.txt} \\
Flowers-102 & 102 & Official train/validation/test split \\
GTSRB & 43 & Official train/test split \\
ImageNet-1K & 1000 & Official train/validation split \\
\hline
\end{tabular}
\end{table}

\subsection{Models}
We evaluate two modified ResNet architectures, namely \textbf{ResNet-18} and \textbf{ResNet-50}.
For both models, the standard ImageNet-style stem is replaced with a CIFAR-style stem, i.e., a \(3\times3\) convolution with stride 1 and no max-pooling.
A global average pooling layer and a single linear classifier are appended on top of the backbone.

\subsection{Input Preprocessing and Augmentation}
Input preprocessing is adjusted for each dataset.
CIFAR-10 and CIFAR-100 are used at their native resolution of \(32\times32\), while higher-resolution datasets are resized to the input resolution specified for the corresponding experiment.
During training, data augmentation is applied in a dataset-dependent manner; for evaluation, no stochastic augmentation is used.
All inputs are normalized with the mean and standard deviation of the corresponding dataset.

\subsection{Training Objectives}
We consider three supervised objectives:
\begin{equation}
\mathcal{L}_{\mathrm{CE}} = -\frac{1}{N}\sum_{i=1}^{N}\log p_{\theta}(y_i \mid x_i),
\end{equation}
\begin{equation}
\mathcal{L}_{\mathrm{CE+L2}} = \mathcal{L}_{\mathrm{CE}} + \frac{\lambda}{2}\lVert \theta \rVert_2^2,
\end{equation}
and
\begin{equation}
\mathcal{L}_{\mathrm{FNG\text{-}CE}}
=
\mathcal{L}_{\mathrm{CE}}
+
\lambda_1 \frac{1}{N}\sum_{i=1}^{N}\left(\|f_i\|_2-c\right)^2
+
\lambda_2
\left\|
\frac{1}{N}\sum_{i=1}^{N}(f_i-\mu)(f_i-\mu)^\top-I
\right\|_F^2
+
\lambda_3 \|W\|_2^2,
\label{eq:fngce_exp_final}
\end{equation}
where \(f_i\) denotes the feature representation of sample \(x_i\), \(\mu\) is the batch mean of features, \(I\) is the identity matrix, and \(W\) denotes the classifier weight matrix. In our default setup, \(\lambda=10^{-4}\) for CE+L2 and \(\lambda=0\) for plain CE.

% \subsection{Optimization and Schedule}
% Unless otherwise specified, models are trained for 50 epochs with Adam optimizer, learning rate \(3\times10^{-4}\), and batch size 256.
% No learning-rate scheduler or early stopping is used.
% We evaluate on the test split at every epoch and report the final-epoch test accuracy.
% Only the final checkpoint is stored.

% \subsection{Evaluation Protocol}
% For each loss type (\(\mathrm{CE}\), \(\mathrm{CE+L2}\)), we run experiments on all six datasets with both backbones, yielding \(2 \times 6 \times 2 = 24\) runs in total.
% Runs are executed in a fixed order: all datasets with ResNet-18 first, then all datasets with ResNet-50.
% Per-run metrics are exported to structured logs (\texttt{JSON}) and aggregated into summary tables (\texttt{CSV}/\texttt{Markdown}).

\subsection{Evaluation Protocol}
We evaluate \(\mathrm{CE}\), \(\mathrm{CE+L2}\), and \(\mathrm{FNG\text{-}CE}\) on all seven datasets with both ResNet-18 and ResNet-50.
For each dataset-backbone pair, the three objectives are trained and evaluated under identical data splits, preprocessing, and optimization settings, so that performance differences can be attributed solely to the training objective.
Unless otherwise specified, performance is measured by top-1 accuracy on the official test split; for ImageNet-1K, we report accuracy on the validation split.
Results are presented for every dataset and backbone combination.

\end{document}